\documentclass[letterpaper]{article}
\usepackage[top=1.25in, bottom=1.5in, left=1.5in, right=1.5in]{geometry}

\usepackage{microtype}
\usepackage{graphicx}
\usepackage{subfigure}
\usepackage{booktabs}
\usepackage{authblk}
\usepackage{amssymb}
\usepackage{amsmath}
\usepackage{amsfonts}
\usepackage{amsthm}
\usepackage{mathtools}
\usepackage{algorithmic}
\usepackage{algorithm}
\usepackage{natbib}
\usepackage{url}

\graphicspath{{./figures/}}

\newcommand{\argmin}{\operatornamewithlimits{\arg \min}}

\newcommand{\bx}{{\mathbf{x}}}
\newcommand{\by}{{\mathbf{y}}}
\newcommand{\bI}{\mathbf{I}}
\newcommand{\bX}{\mathbf{X}}
\newcommand{\blambda}{{\boldsymbol \lambda}}

\newcommand{\calL}{\mathcal{L}}
\newcommand{\calO}{\mathcal{O}}
\newcommand{\calX}{\mathcal{X}}
\newcommand{\bbR}{\mathbb{R}}
\newcommand{\bbZ}{\mathbb{Z}}

\begin{document}

\title{Practical Bayesian Optimization with Threshold-Guided Marginal Likelihood Maximization}

\author[]{Jungtaek Kim}
\author[]{Seungjin Choi\thanks{S. Choi is currently affiliated with BARO AI.}}
\affil[]{\normalsize Pohang University of Science and Technology, Pohang, Republic of Korea}
\affil[]{\texttt{\{jtkim,seungjin\}@postech.ac.kr}}

\maketitle

\begin{abstract}
We propose a practical Bayesian optimization method
using Gaussian process regression, of which the marginal likelihood 
is maximized where the number of model selection steps is guided by a pre-defined threshold.
Since Bayesian optimization consumes a large portion of its execution time 
in finding the optimal free parameters 
for Gaussian process regression, our simple, but straightforward method is able to mitigate 
the time complexity and speed up the overall Bayesian optimization procedure.
Finally, the experimental results show that our method is effective to reduce the 
execution time in most of cases, with less loss of optimization quality.
\end{abstract}

\section{Introduction\label{sec:intro}}

Bayesian optimization is a global optimization method with an acquisition function, 
which is induced by a surrogate function.
Since in the problem formulation of Bayesian optimization we often assume that an objective function is unknown, 
the acquisition function is optimized instead.
For this principle of Bayesian optimization, 
the selection of surrogate function and acquisition function is a design choice that should be carefully considered,
in order to find a global optimizer quickly.

In general, 
Gaussian process regression~\citep{RasmussenCE2006book} is mainly taken into account as surrogate function for the reason why its complexity and model capacity are sufficiently large and flexible~\citep{SnoekJ2012neurips},
and we usually choose a simple acquisition function such as expected improvement~\citep{MockusJ1978tgo} and Gaussian process upper confidence bound~\citep{SrinivasN2010icml} 
because of their powerful performance~\citep{SnoekJ2012neurips,SuiY2015icml,SpringenbergJT2016neurips,KimJ2018automl}.

However, even though evaluating and analyzing the convergence quality of Bayesian optimization is important, 
its execution time is also significant.
In a single round of Bayesian optimization,
Bayesian optimization with Gaussian process regression spends most of the execution time in two distinct steps:
(i) building a Gaussian process regression model with optimal kernel free parameters 
and (ii) optimizing an acquisition function to find the next query point.

The main interest of this paper is to alleviate a time complexity 
for the issue on Gaussian process regression, assuming a generic condition of Bayesian optimization 
and its circumstance.
To be precise, a model selection step for Gaussian process accounts 
for a large portion of the consumed time, which is originated from matrix inverse operations 
(i.e., the time complexity of which is $\calO(n^3)$ where $n$ is the number of data points).
In this paper we tackle this problem, 
by proposing the method to reduce the time consumed in the model selection step.
Since a surrogate function produces similar outputs as query points are accumulated, 
optimal free parameters of Gaussian process regression have not been changed dramatically.
Thus, in practice we might skip the model selection step where sufficiently enough data points have been observed.
This observation encourage us to suggest our method.
From now, we introduce backgrounds and main idea.

\section{Background\label{sec:back}}

In this section, we briefly introduce Gaussian process regression 
and model selection techniques for Gaussian process regression, which are discussed in this work.

\subsection{Gaussian Process Regression\label{subsec:gpr}}

Gaussian process regression is one of popular Bayesian nonparametric regression methods, 
which provides a function estimate and its uncertainty estimate.
The outputs of Gaussian process regression are used to balance exploration and exploitation 
in the perspective of Bayesian optimization.
To compute a posterior mean function and a posterior variance function, 
we need to set appropriate kernel free parameters for Gaussian process regression, 
using model selection which is described in Section~\ref{subsec:model}.
Due to a space limit, we omit the detailed explanation of Gaussian process regression; 
See \citet{RasmussenCE2006book} for the details.

\subsection{Model Selection of Gaussian Process Regression\label{subsec:model}}

We choose one of two popular model selection techniques of Gaussian process regression:
(i) marginal likelihood maximization (MLM)
and (ii) leave-one-out cross-validation (LOO-CV),
to find an optimal model that expresses the given dataset well.
The marginal likelihood over function values $\by \in \bbR^{n}$ 
conditioned on inputs $\bX \in \bbR^{n \times d}$ and kernel free parameters $\blambda$ (in this paper $\blambda \in \bbR^{d + 1}$, but it is differed as a type of kernel) is
\begin{align}
	\calL_{\textrm{ML}} &=
	\log p(\by | \bX, \blambda) \nonumber\\
	&= -\frac{1}{2} \by^\top (K(\bX, \bX) + \sigma_n^2 \bI)^{-1} \by - \frac{1}{2} \log \det (K(\bX, \bX) + \sigma_n^2 \bI) - \frac{n}{2} \log 2 \pi,
	\label{eqn:marginal_l}
\end{align}
where $K(\bX, \bX) \in \bbR^{n \times n}$ is a covariance function that maps from two matrices to all pairwise comparisons in two matrices, and $\sigma_n^2$ is an observation noise level.
\eqref{eqn:marginal_l} is maximized to choose optimal kernel free parameters in the MLM.

The leave-one-out log predictive probability (a.k.a. log pseudo-likelihood) is 
\begin{equation}
	\calL_{\textrm{LOO}} = 
	\sum_{i = 1}^n \log p(y_i | \bX, \by_{-i}, \blambda),
	\label{eqn:loo}
\end{equation}
where
\begin{equation}
	\log p(y_i | \bX, \by_{-i}, \blambda) = - \frac{1}{2} \log \sigma_i^2 - \frac{(y_i - \mu_i)^2}{2 \sigma_i^2} - \frac{1}{2} \log 2 \pi.
\end{equation}
Note that $\by_{-i} \in \bbR^{n - 1}$ is all function values except index $i$ and $\mu_i / \sigma_i^2$ are posterior mean/variance functions.
Since we can preemptively compute the inverse of $K(\bX, \bX) + \sigma_n^2 \bI$, 
the time complexity of \eqref{eqn:loo} is almost similar to \eqref{eqn:marginal_l}.

\section{Main Algorithm\label{sec:main}}

We propose a practical Bayesian optimization framework with threshold-guided MLM,
which leads to speed up Bayesian optimization.
As mentioned above, we assume that optimal kernel free parameters of Gaussian process regression are not significantly changed as Bayesian optimization procedure is iterated.
To measure whether the model is converged or not, we can consider some metrics 
such as marginal likelihood and pseudo-likelihood.
However they are not appropriate to measure the discrepancy between two models we attempt to compare, 
because they are computed from different data points and their corresponding observations 
(in practice two models are the models at iteration $t$ and iteration $t+1$).
For this reason, we simply compare the free parameters at iteration $t$ with the free parameters at the subsequent iteration, by computing $l_2$ distance between them.

\begin{algorithm}[t]
	\caption{Practical Bayesian Optimization with Threshold-Guided MLM}
	\label{alg:main}
	\begin{algorithmic}[1]
		\REQUIRE Function domain $\calX \in \bbR^d$ where $d$ is a dimensionality of domain, the number of initial points $t \in \bbZ > 0$, iteration budget $\tau \in \bbZ > 0$, and a threshold for validating convergence $\rho \in \bbR > 0$
		\ENSURE Point $\bx^\dagger$ that has shown the best observation.
		\STATE Sample $t$ initial points randomly $\{ \bx_i \}_{i = 1}^t$ where $\bx \in \calX$.
		\STATE Observe $t$ points from $y = f(\bx) + \epsilon$ where $\epsilon$ is an observation noise, $\{ y_i \}_{i = 1}^t$.
		\FOR {$j = t + 1, \ldots, \tau$}
			\IF {$j > t + 2$ \textbf{and} $\| \blambda_{j - t - 1}^* - \blambda_{j - t - 2}^* \|_2 < \rho$}
				\STATE Build a surrogate model via Gaussian process regression with the previous kernel free parameters $\blambda_{j - t - 1}^*$, $\hat{f}(\cdot | \{ (\bx_i, y_i) \}_{i = 1}^{j - 1}, \blambda_{j - t - 1}^*)$.
				\STATE Consider $\blambda_{j - t}^*$ as $\blambda_{j - t - 1}^*$.
			\ELSE
				\STATE Build a surrogate model via Gaussian process regression with MLM over kernel free parameters, $\hat{f}(\cdot | \{ (\bx_i, y_i) \}_{i = 1}^{j - 1}, \blambda_{j - t}^*)$.
			\ENDIF
			\STATE Keep the kernel free parameters $\{\blambda^*_i\}_{i = 1}^{j - t}$.
			\STATE Query a point to sample $\bx^*$, optimizing an acquisition function $a(\cdot | \{ (\bx_i, y_i) \}_{i = 1}^{j - 1})$.
			\STATE Update historical points and their associated observations $\{ (\bx_i, y_i) \}_{i = 1}^{j}$.
		\ENDFOR
		\STATE \textbf{return} the point that has shown the best observation $\bx^\dagger = \argmin_{(\bx, y) \in \{ (\bx_i, y_i) \}_{i = 1}^\tau} y$.
	\end{algorithmic}
\end{algorithm}

Our method is described in Algorithm~\ref{alg:main}.
First of all, we provide four inputs: (i) a function domain $\calX \in \bbR^d$ where $d$ is a domain dimensionality, 
(ii) the number of initial points $t \in \bbZ > 0$, 
(iii) iteration budget $\tau \in \bbZ > 0$, 
and (iv) a convergence threshold $\rho \in \bbR > 0$.
Our method basically follows an ordinary Bayesian optimization procedure.
After it finds $t$ initial points and their observations, iterate querying and observing steps $\tau$ times, 
as shown in Line 4 to 12 of Algorithm~\ref{alg:main}.
The main difference between an ordinary Bayesian optimization and our method is presented in Line 4 to 10 of Algorithm~\ref{alg:main}.
We assess the distance between the last two kernel free parameters 
after obtaining at least two optimal free parameters, as shown in Line 4.
If the distance is larger than $\rho$, new optimal free parameters are found by maximizing marginal likelihood.
Otherwise, the previous kernel free parameters are used to build a surrogate function.
Due to logical flow and space limit, we omit the detailed explanation of ordinary Bayesian optimization.
See the details of Bayesian optimization in \citep{BrochuE2010arxiv,ShahriariB2016procieee,FrazierPI2018arxiv}.

Although it is a practical and simple algorithm to find a global optimum of black-box function,
it shows fair performance according to our experimental results.
Moreover, if we think over the implication of free parameters for Gaussian process, 
this approach is intuitively reasonable.
To show our method is effective, in the next section we demonstrate the results on nine benchmark functions and two real-world problems.
Furthermore, we will introduce the future directions to develop this idea to more theoretical and more sophisticated method.

\section{Experiments\label{sec:exps}}

\begin{figure}[t!]
	\centering
	\subfigure[Beale / 100 iters]{
		\includegraphics[width=0.31\linewidth, keepaspectratio]{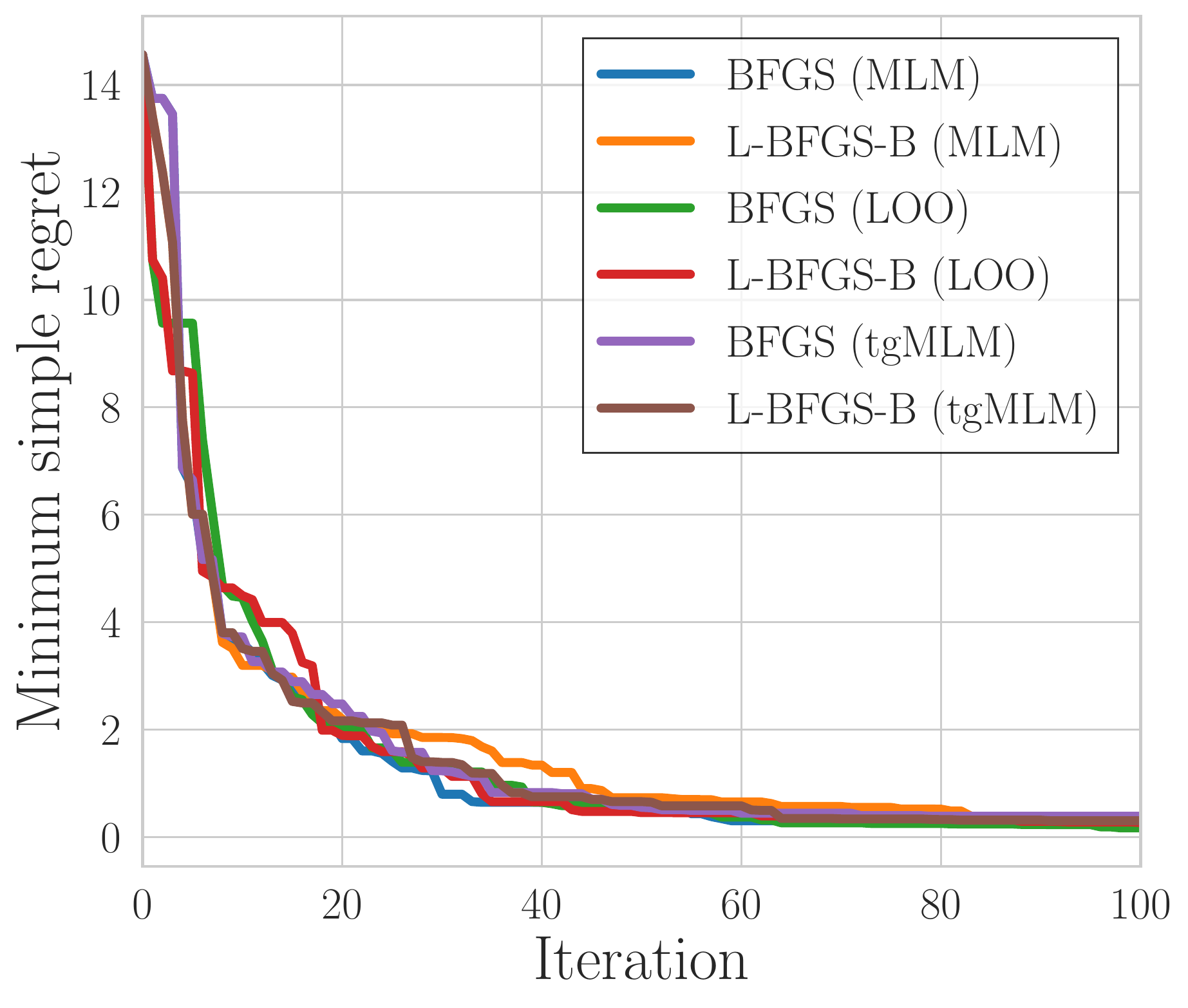}
	}
	\subfigure[Bohachevsky / 100 iters]{
		\includegraphics[width=0.31\linewidth, keepaspectratio]{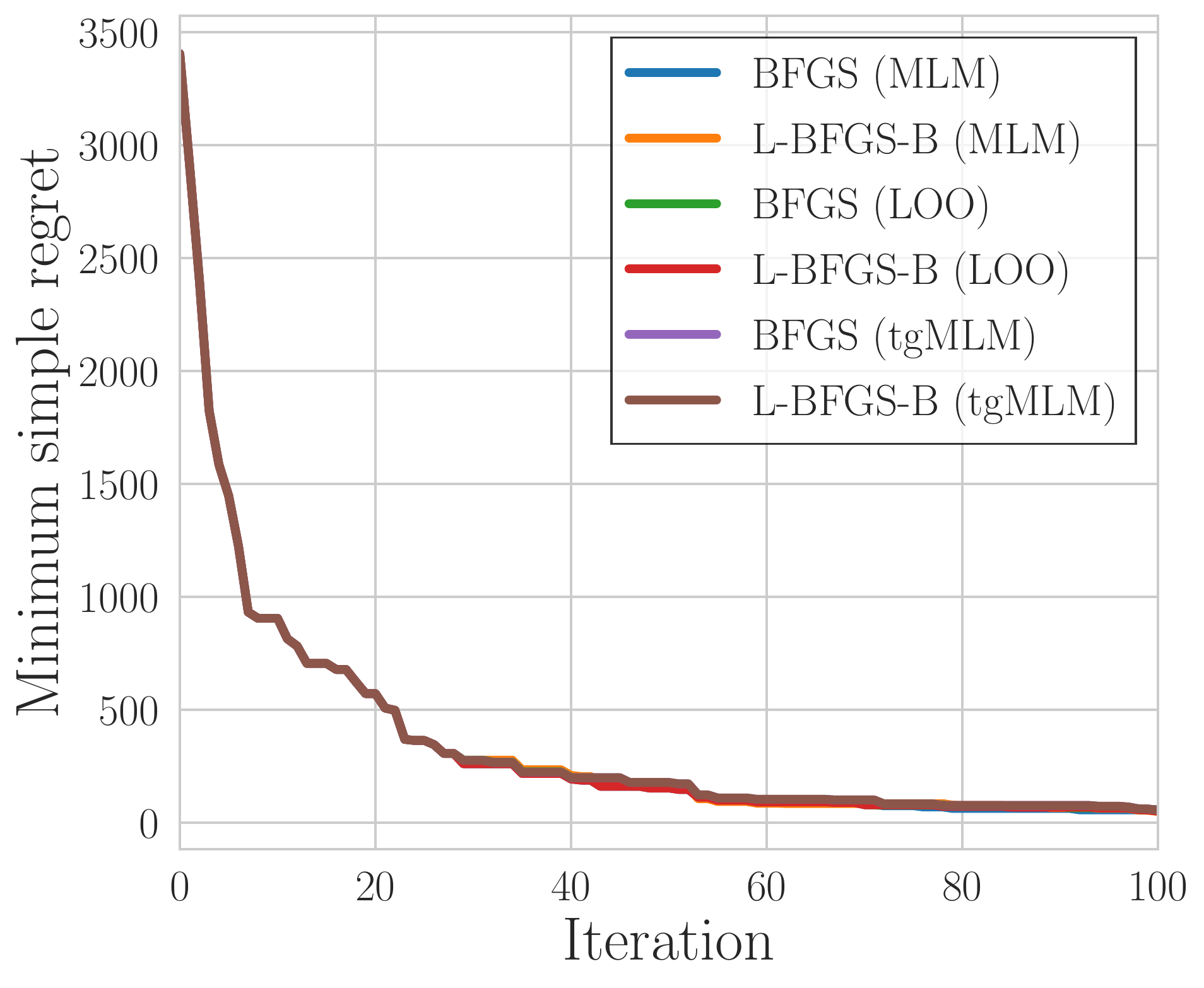}
	}
	\subfigure[Branin / 50 iters]{
		\includegraphics[width=0.31\linewidth, keepaspectratio]{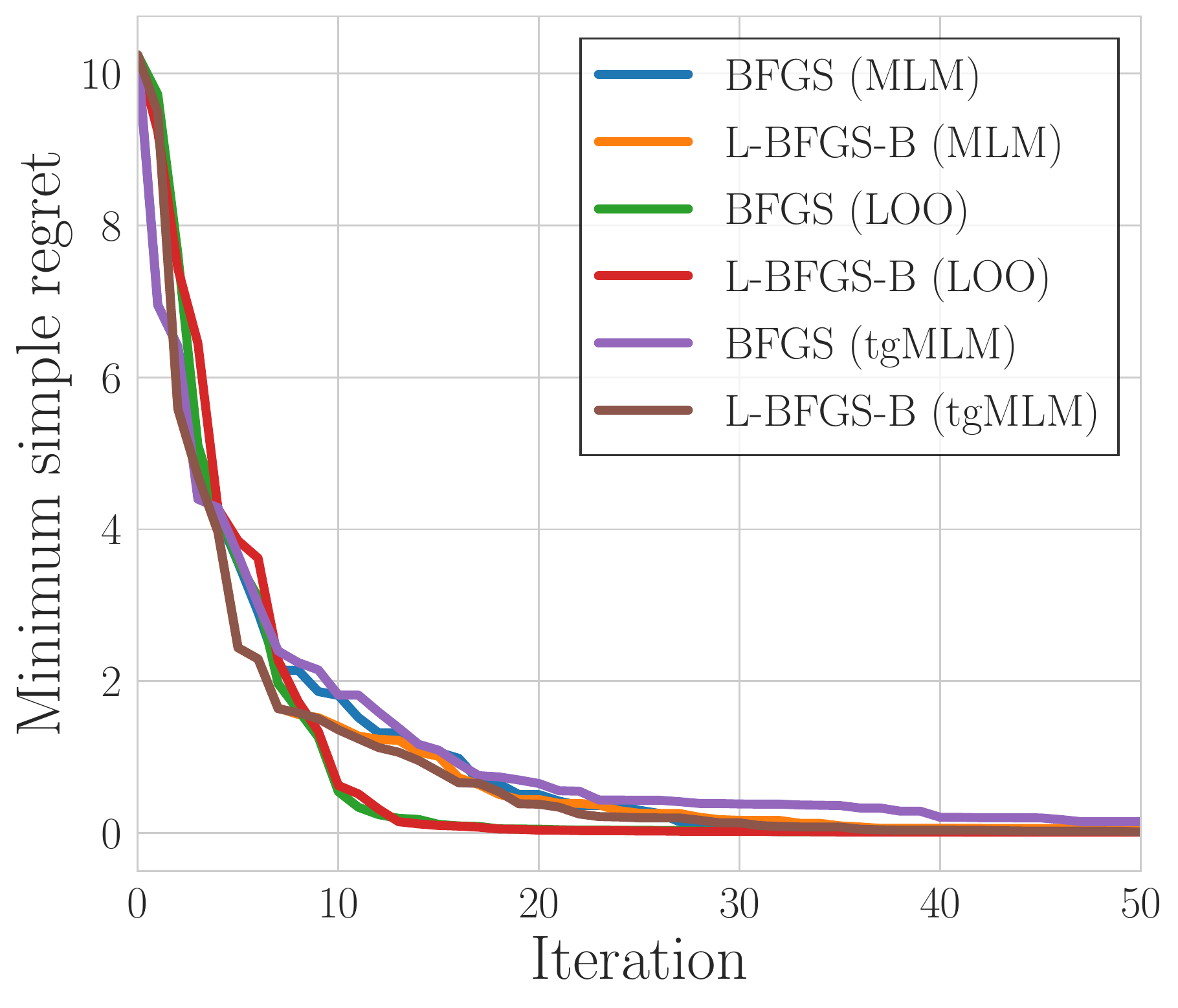}
	}
	\subfigure[Eggholder / 250 iters]{
		\includegraphics[width=0.31\linewidth, keepaspectratio]{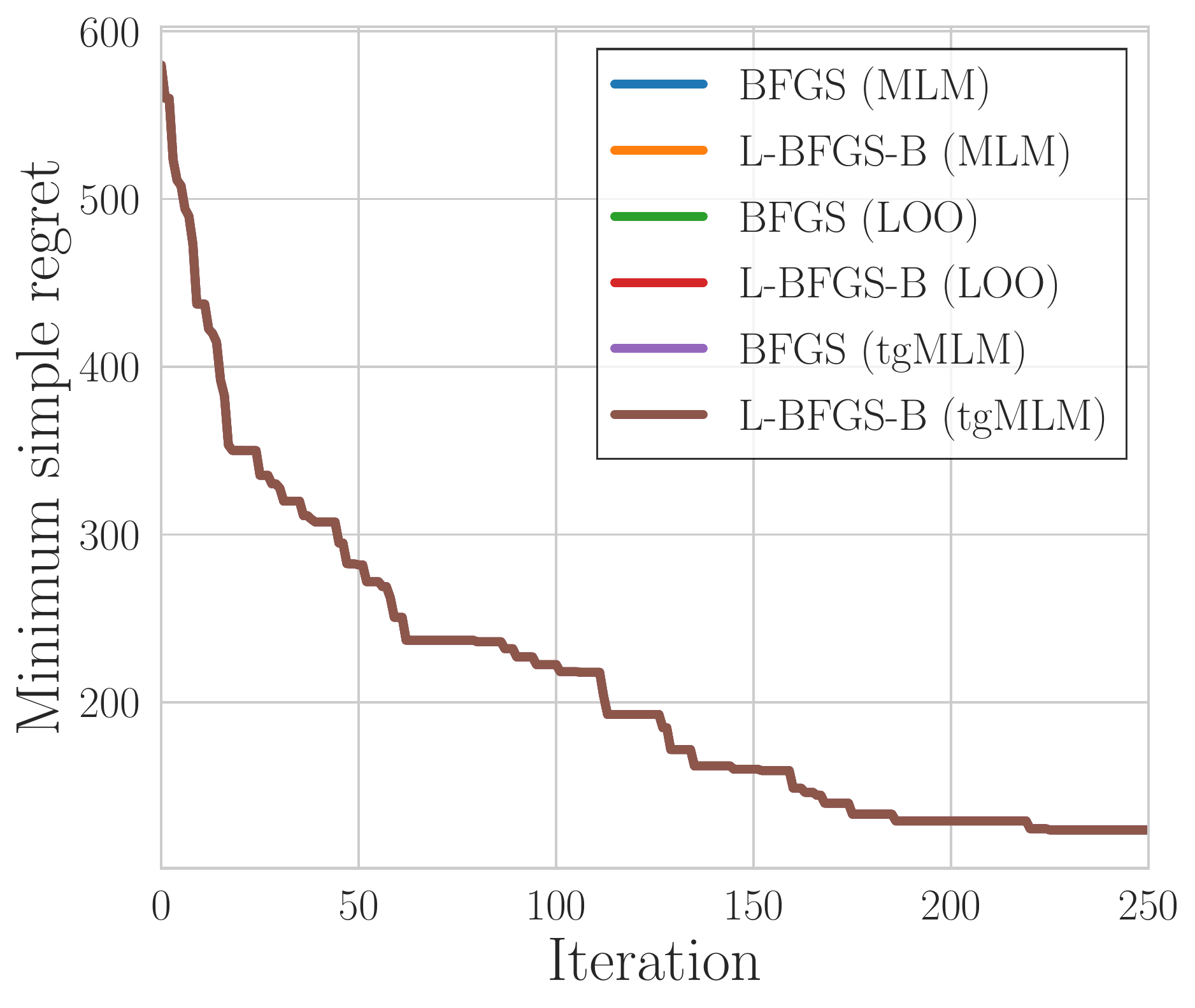}
	}
	\subfigure[Goldstein-Price / 50 iters]{
		\includegraphics[width=0.31\linewidth, keepaspectratio]{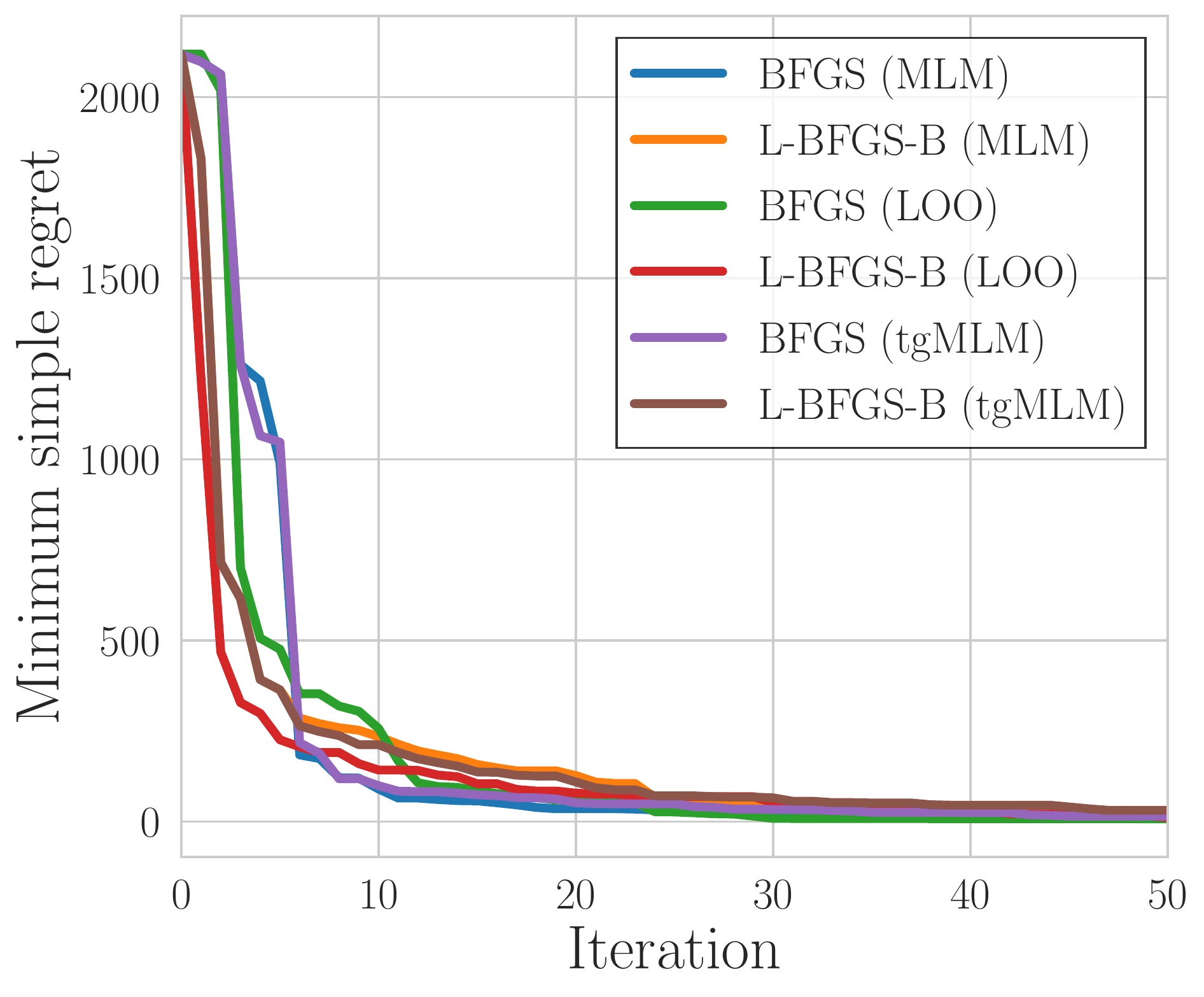}
	}
	\subfigure[Hartmann6D / 250 iters]{
		\includegraphics[width=0.31\linewidth, keepaspectratio]{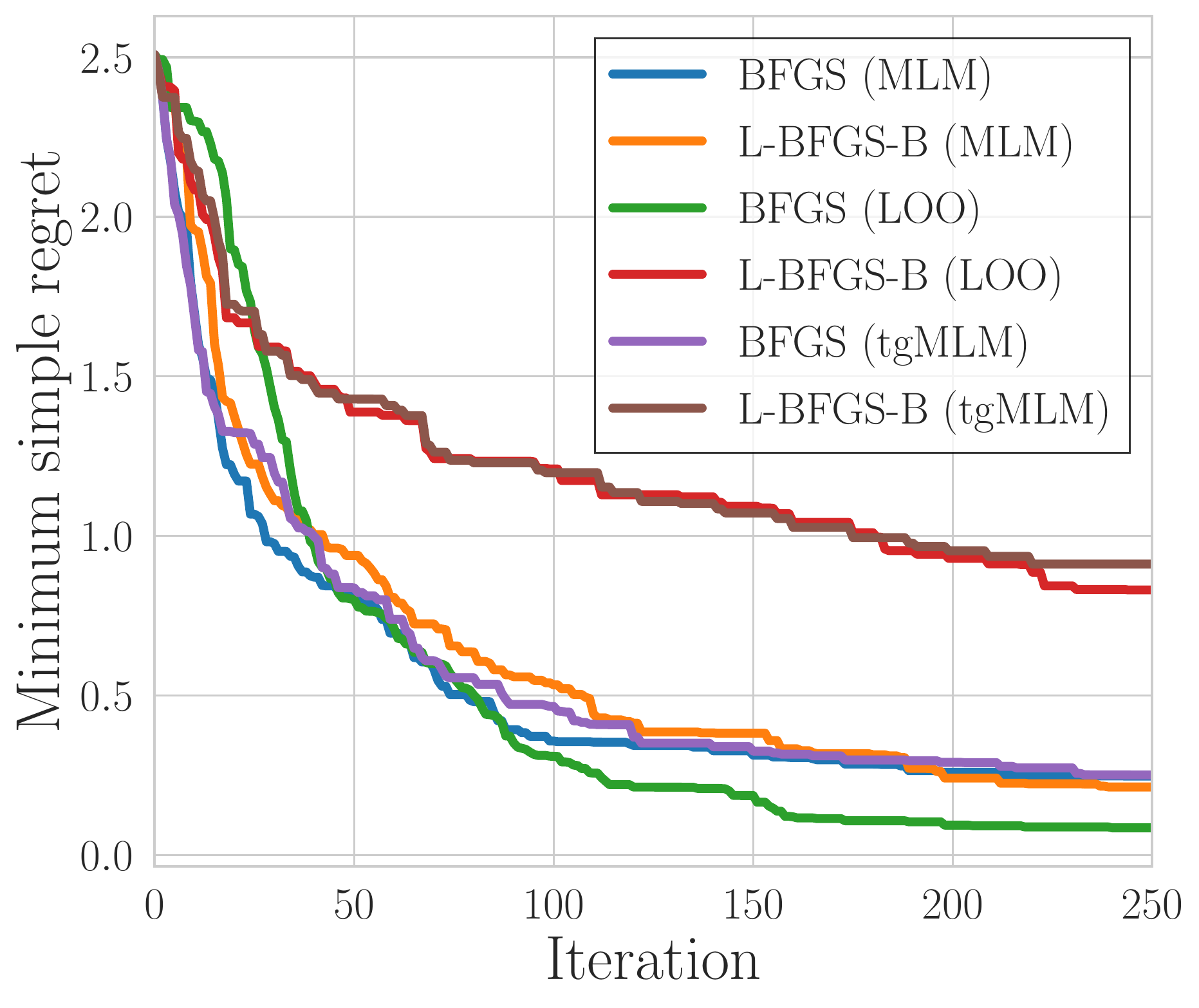}
	}
	\subfigure[Holder Table / 100 iters]{
		\includegraphics[width=0.31\linewidth, keepaspectratio]{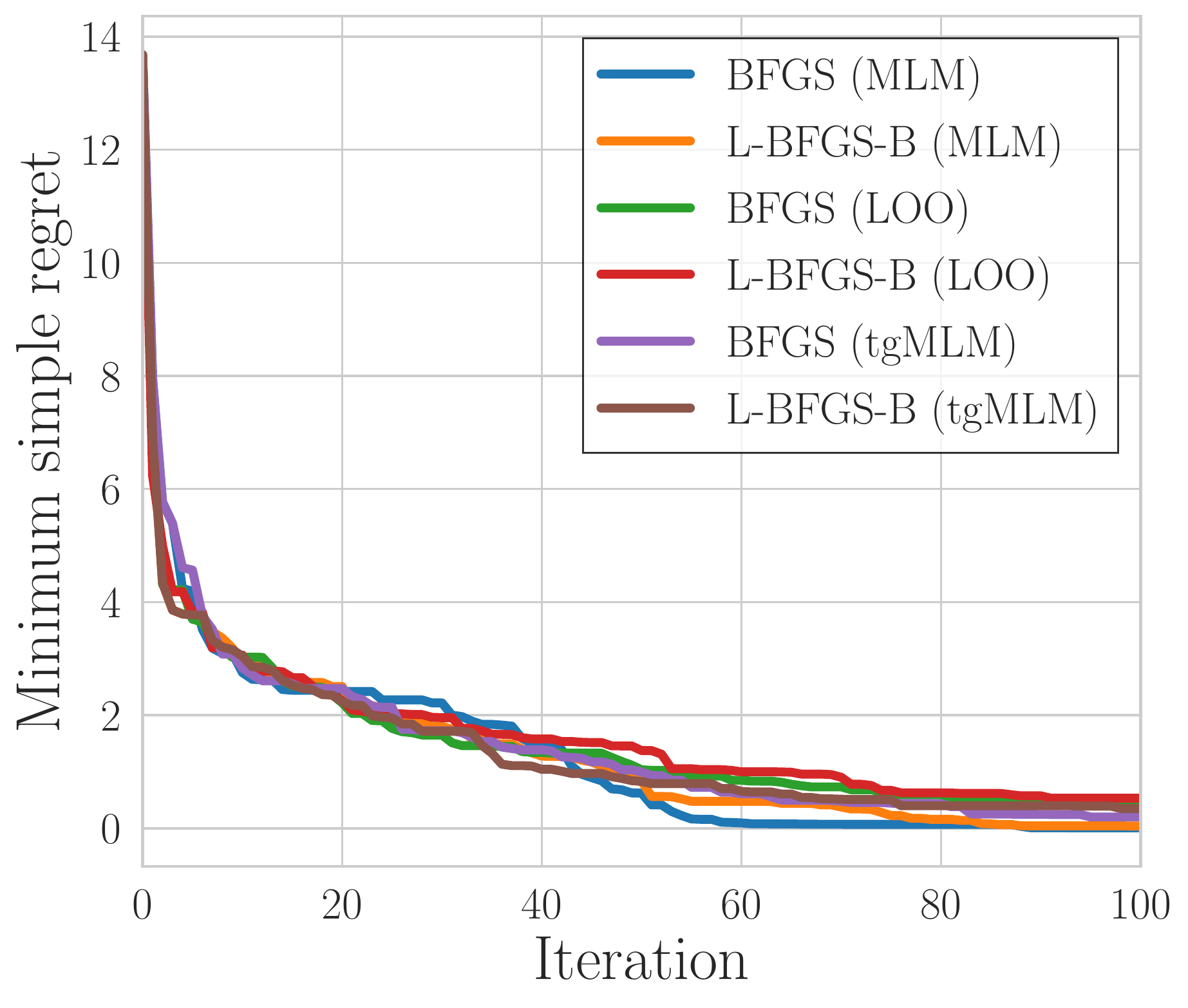}
	}
	\subfigure[Rosenbrock / 100 iters]{
		\includegraphics[width=0.31\linewidth, keepaspectratio]{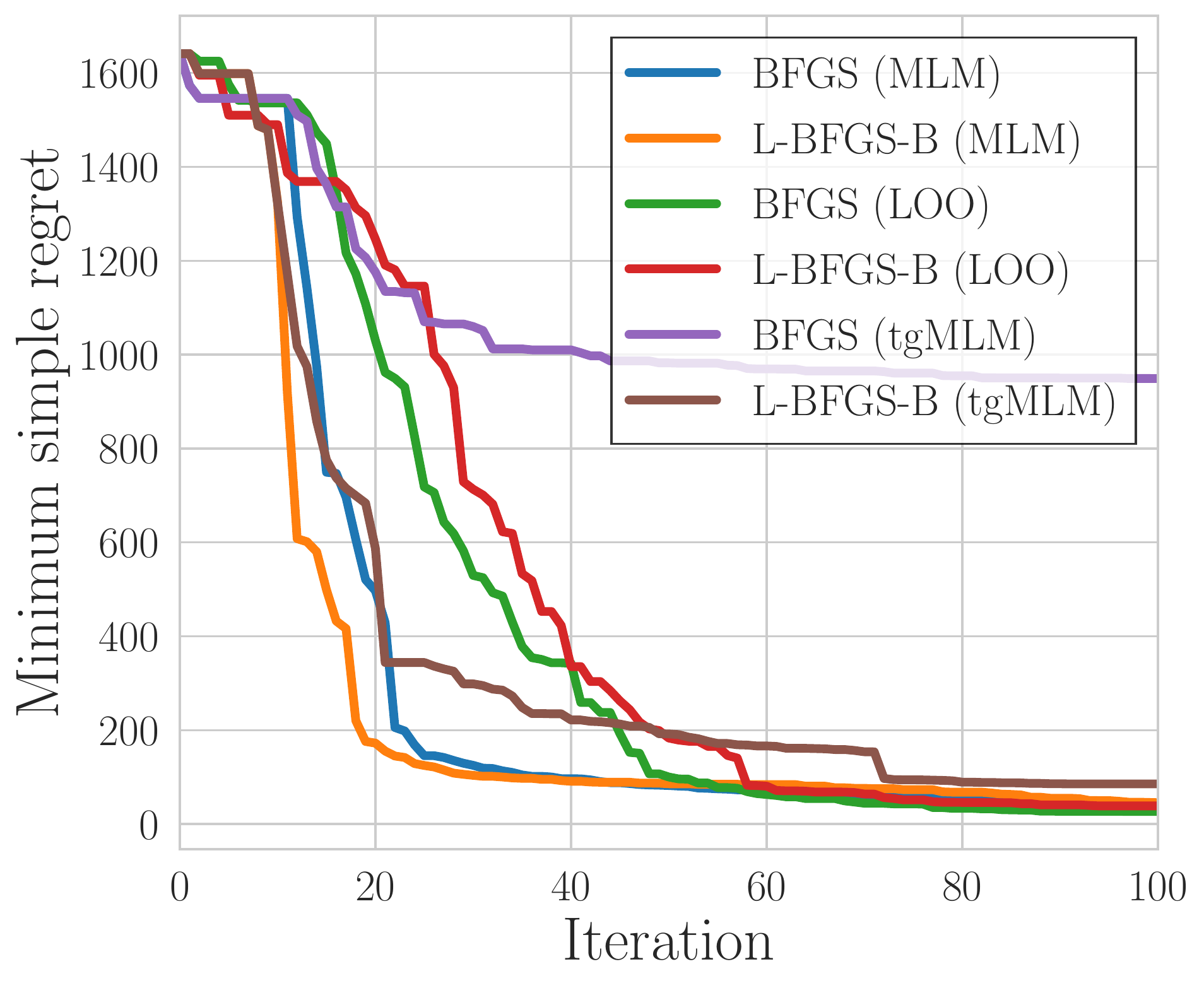}
	}
	\subfigure[Six-hump Camel / 100 iters]{
		\includegraphics[width=0.31\linewidth, keepaspectratio]{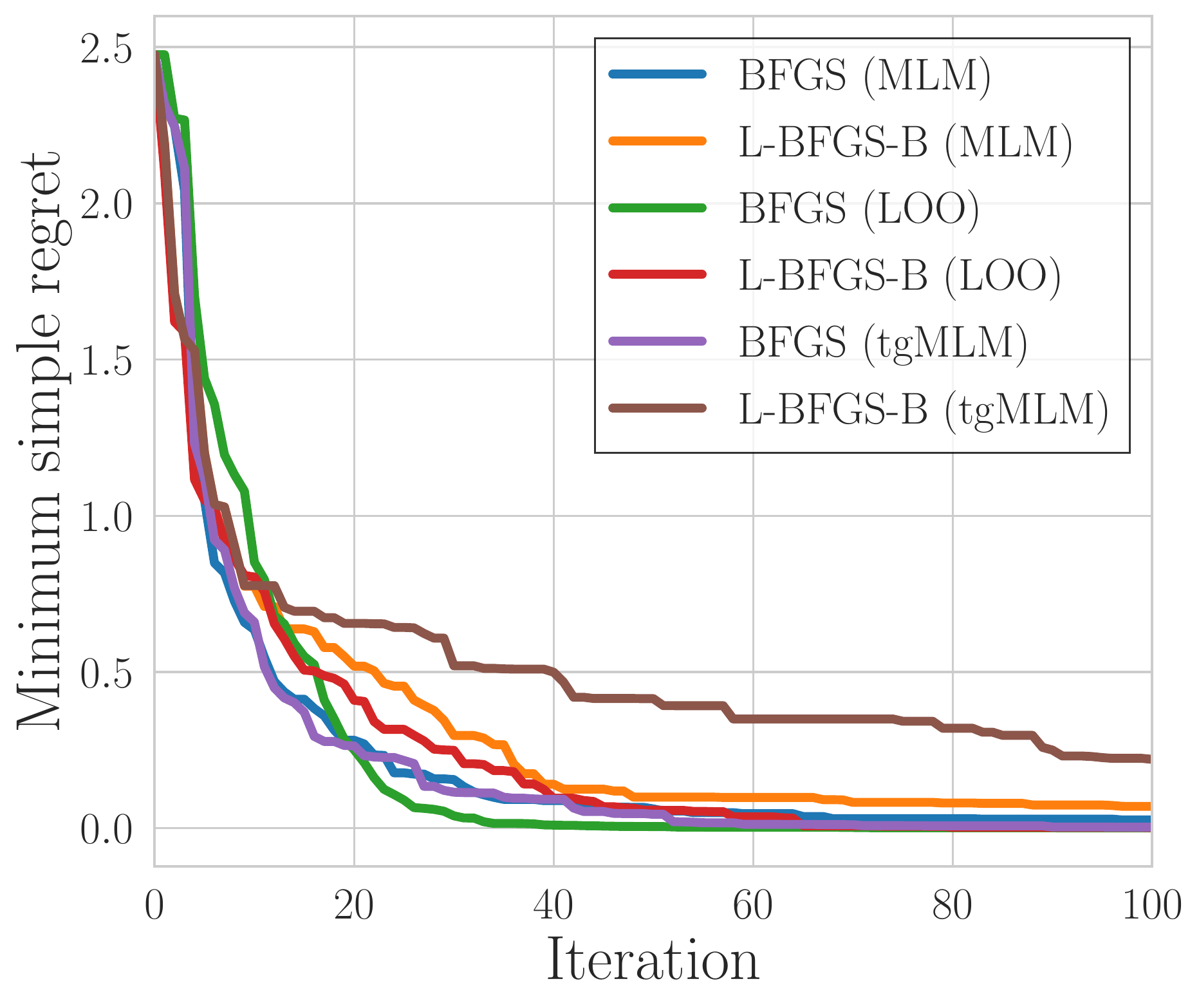}
	}
	\subfigure[RW-1 / 100 iters]{
		\includegraphics[width=0.31\linewidth, keepaspectratio]{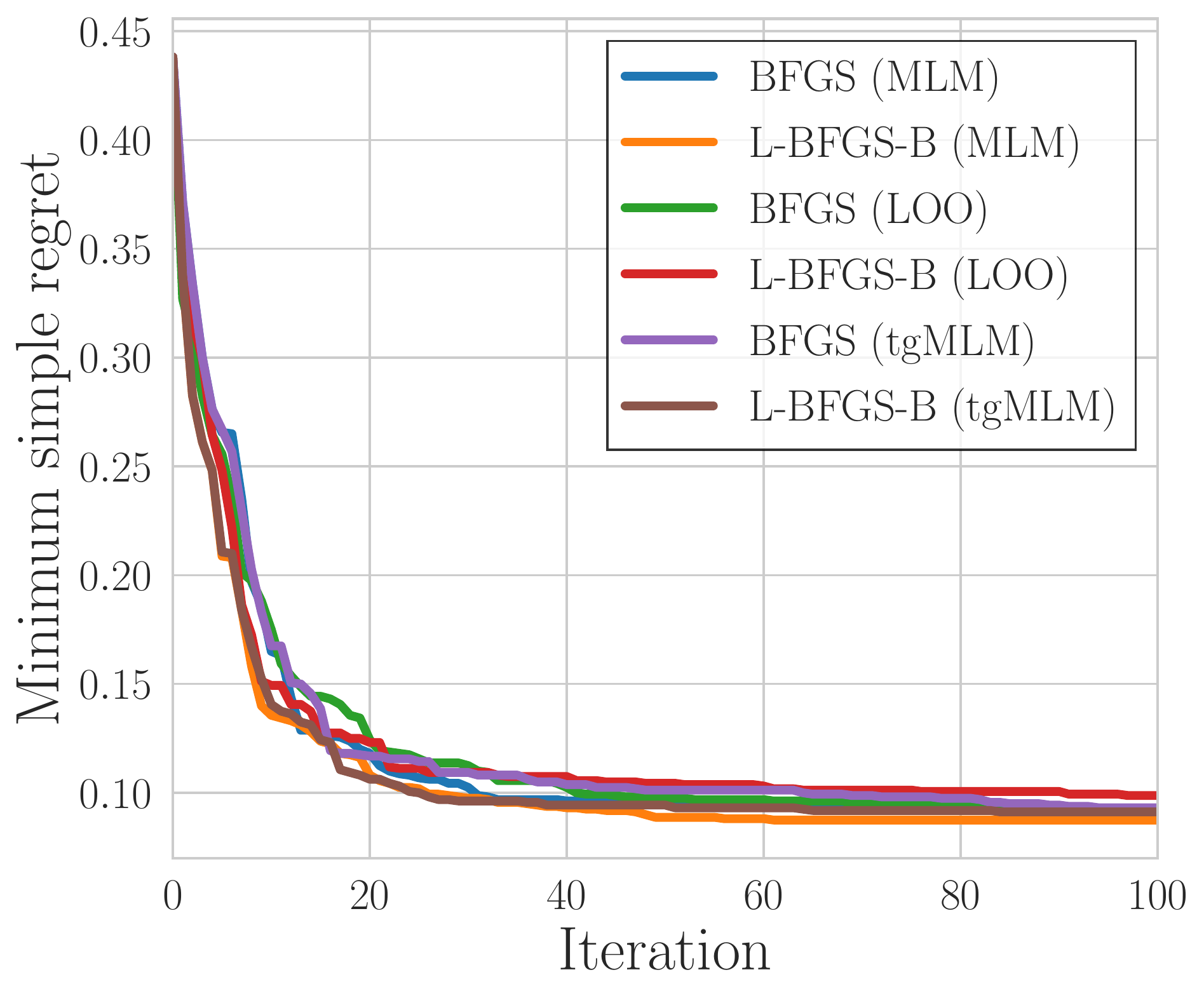}
	}
	\subfigure[RW-2 / 100 iters]{
		\includegraphics[width=0.31\linewidth, keepaspectratio]{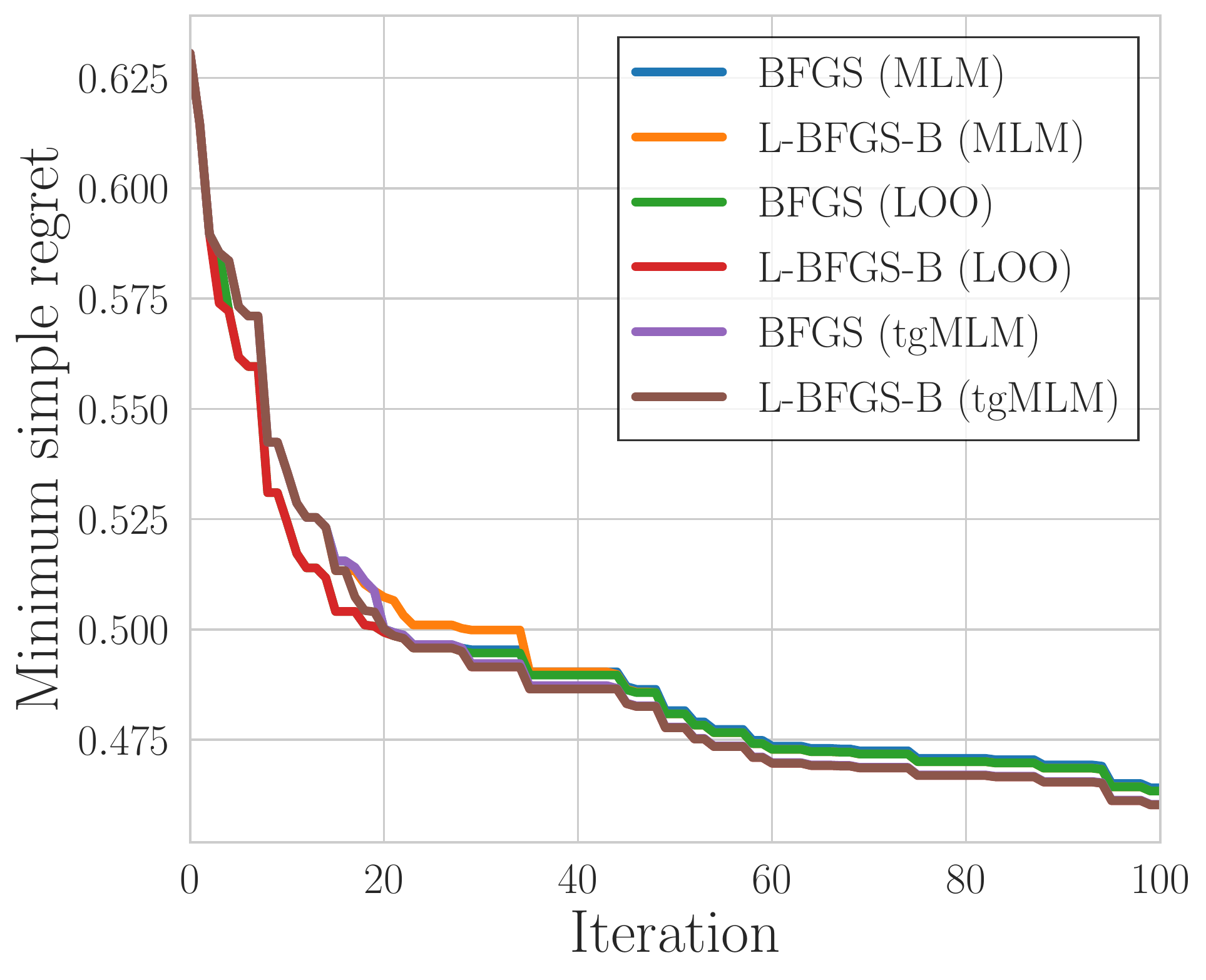}
	}
	\caption{Results on Bayesian optimization of nine benchmark functions and two real-world problems.
	LOO and tgMLM indicate LOO-CV and threshold-guided MLM, respectively.
	All experiments are repeated 20 times.}
	\label{fig:all_exps_1}
\end{figure}

\begin{figure}[t!]
	\centering
	\subfigure[Beale / 100 iters]{
		\includegraphics[width=0.31\linewidth, keepaspectratio]{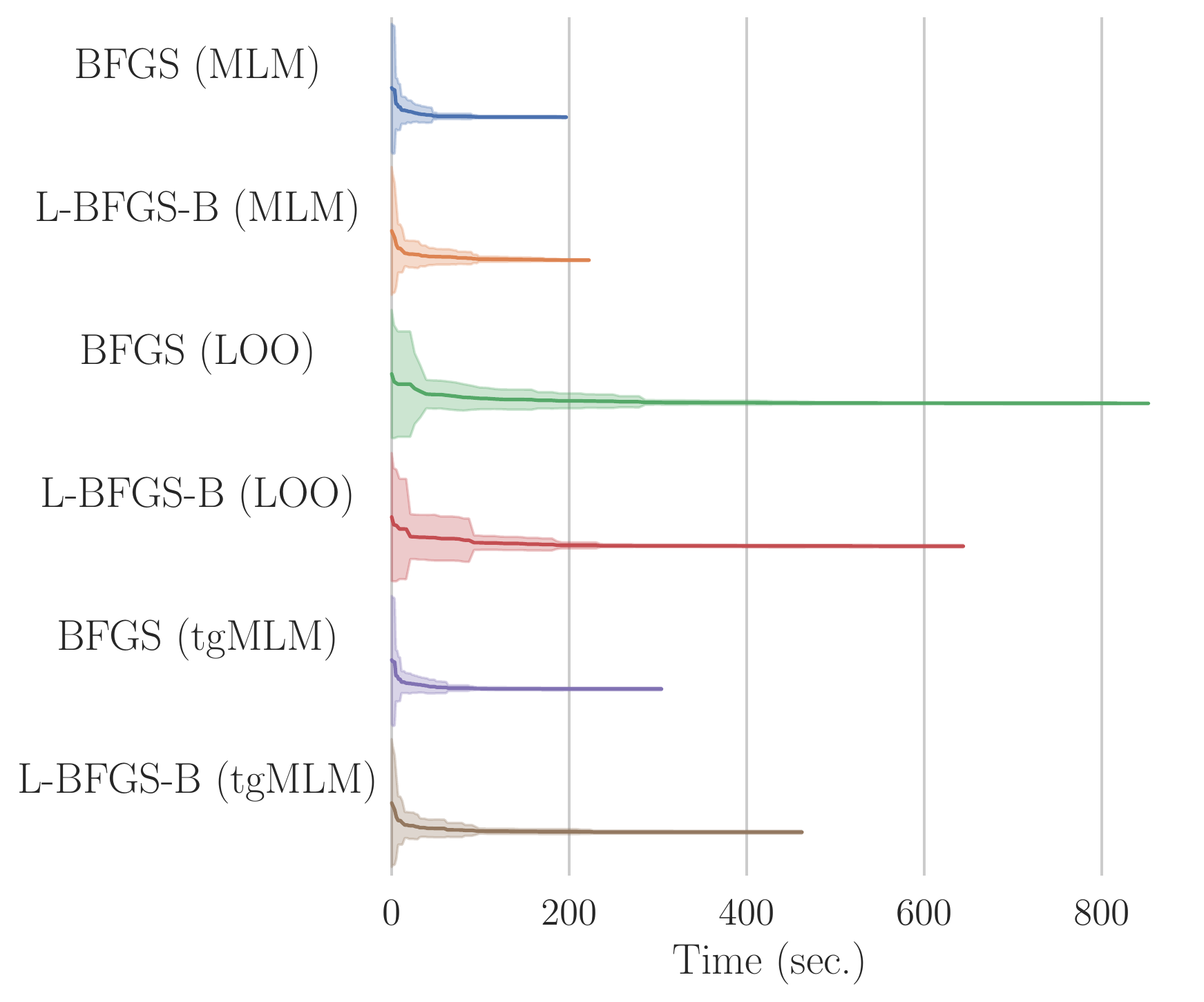}
	}
	\subfigure[Bohachevsky / 100 iters]{
		\includegraphics[width=0.31\linewidth, keepaspectratio]{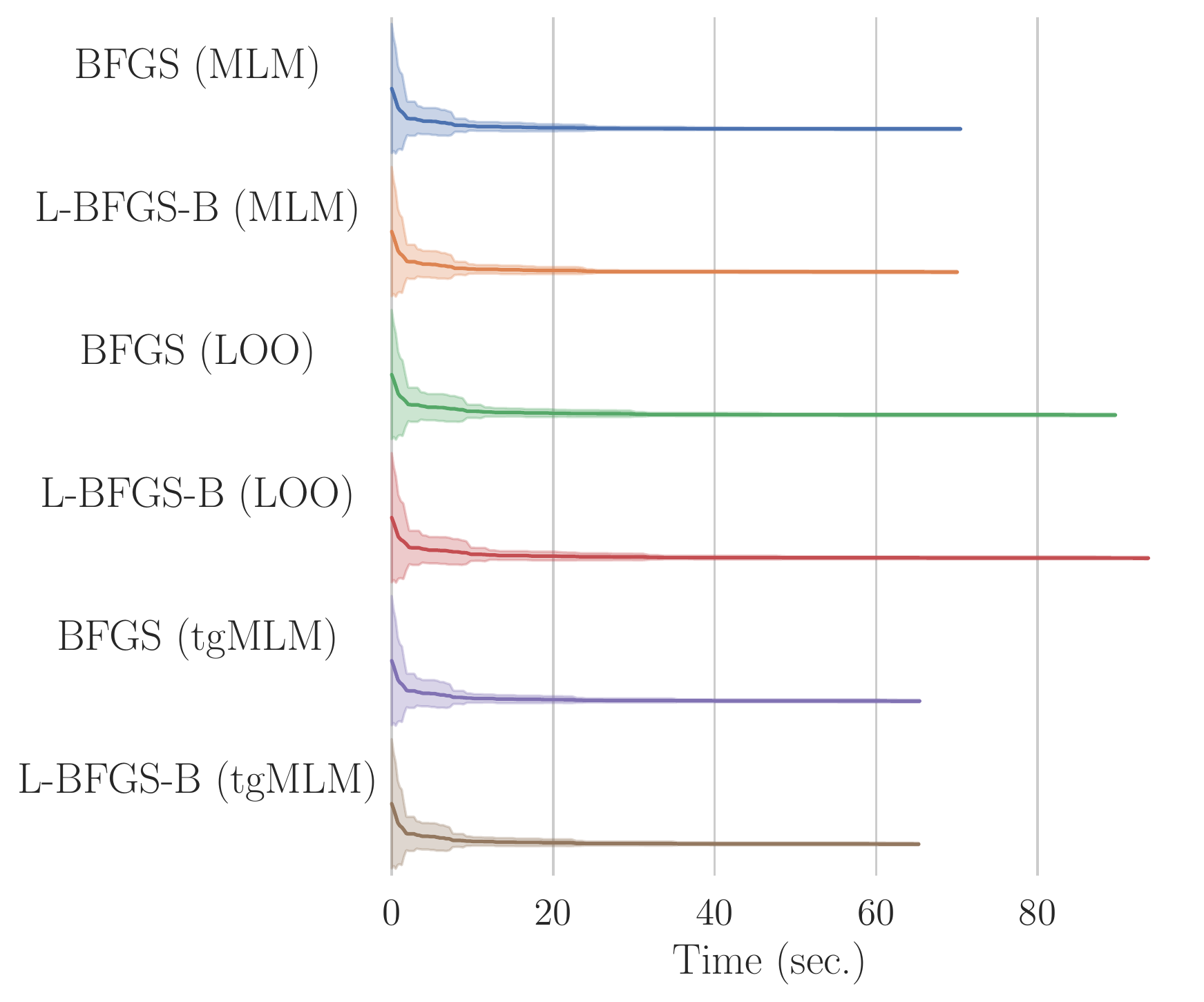}
	}
	\subfigure[Branin / 50 iters]{
		\includegraphics[width=0.31\linewidth, keepaspectratio]{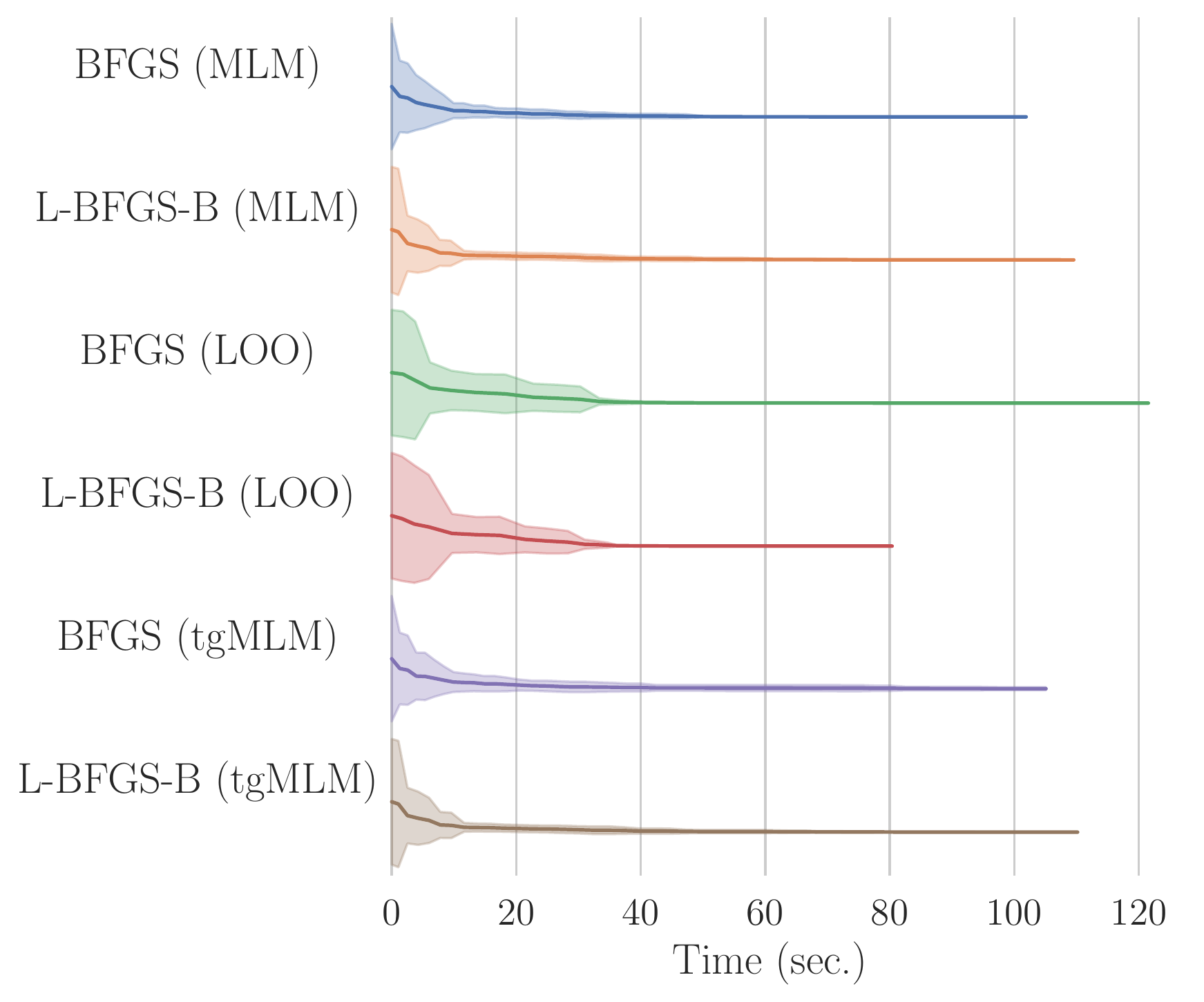}
	}
	\subfigure[Eggholder / 250 iters]{
		\includegraphics[width=0.31\linewidth, keepaspectratio]{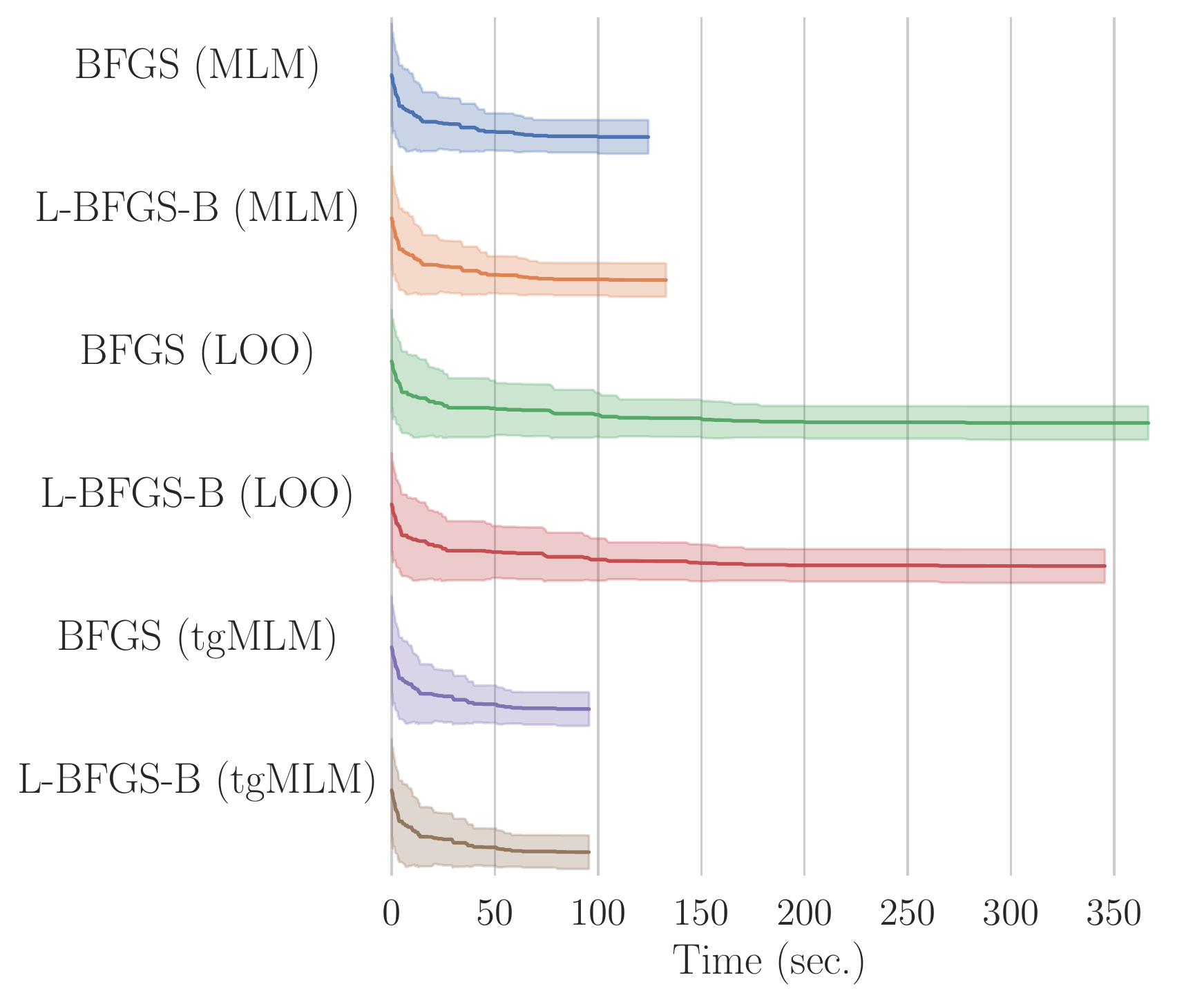}
	}
	\subfigure[Goldstein-Price / 50 iters]{
		\includegraphics[width=0.31\linewidth, keepaspectratio]{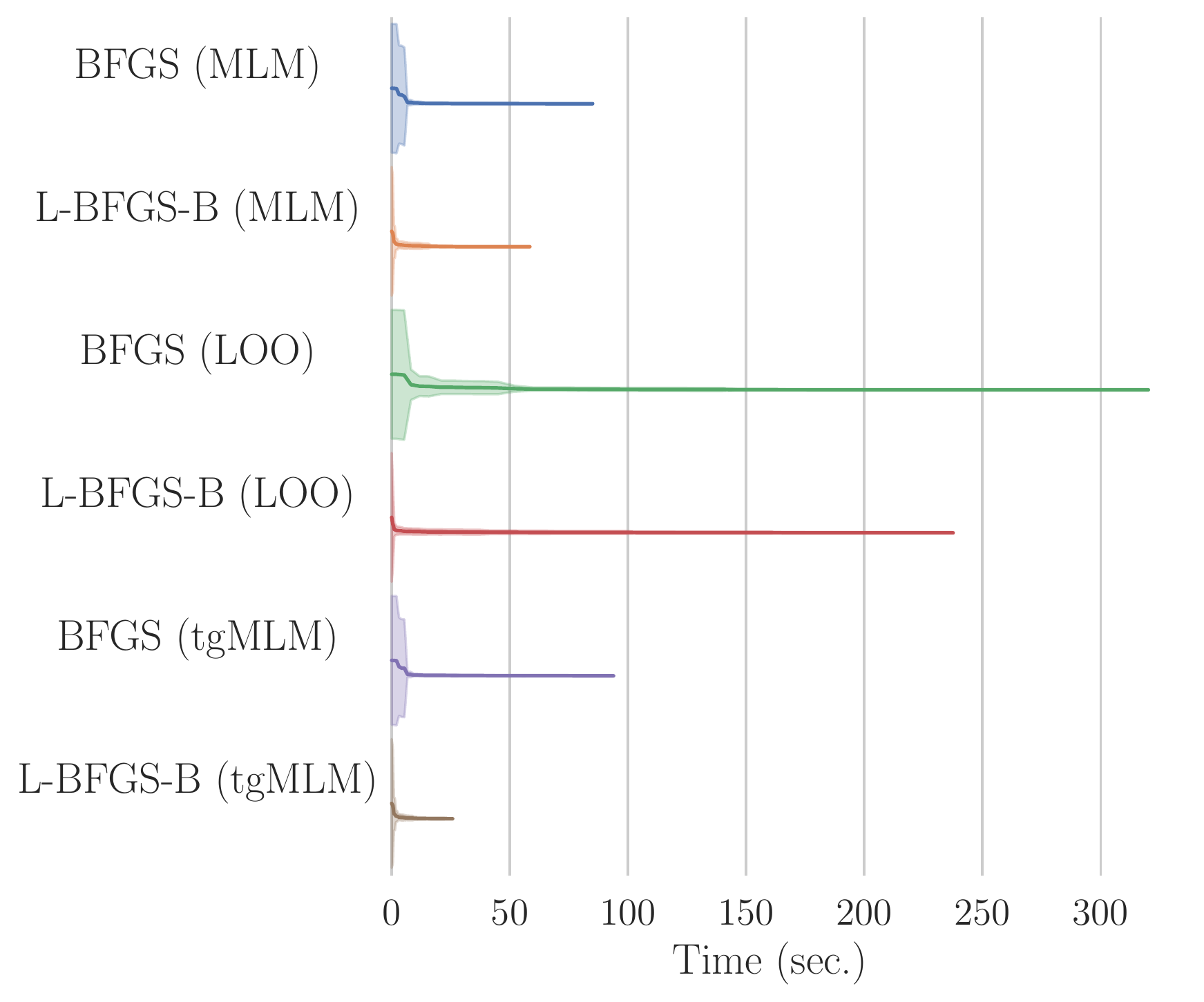}
	}
	\subfigure[Hartmann6D / 250 iters]{
		\includegraphics[width=0.31\linewidth, keepaspectratio]{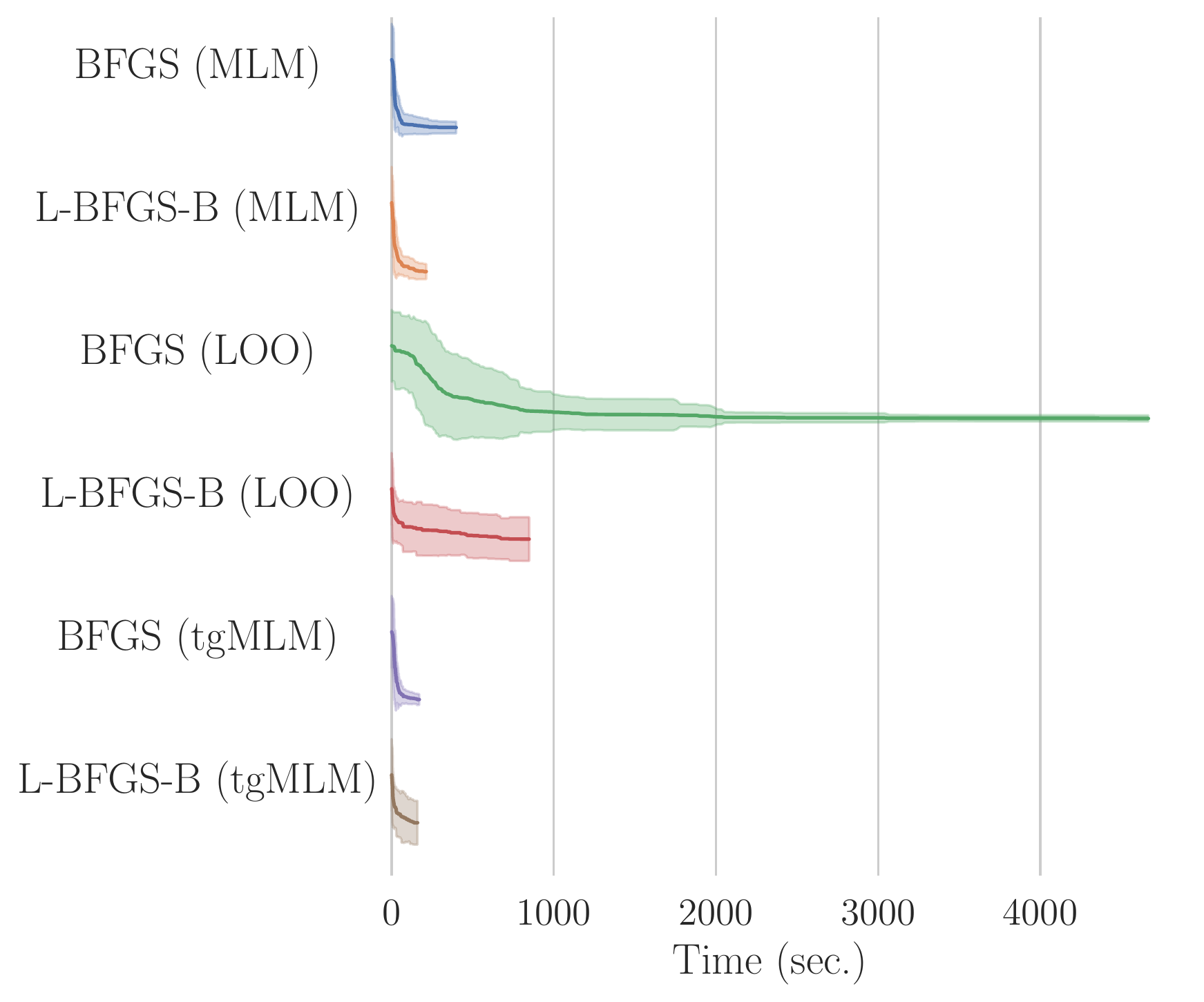}
	}
	\subfigure[Holder Table / 100 iters]{
		\includegraphics[width=0.31\linewidth, keepaspectratio]{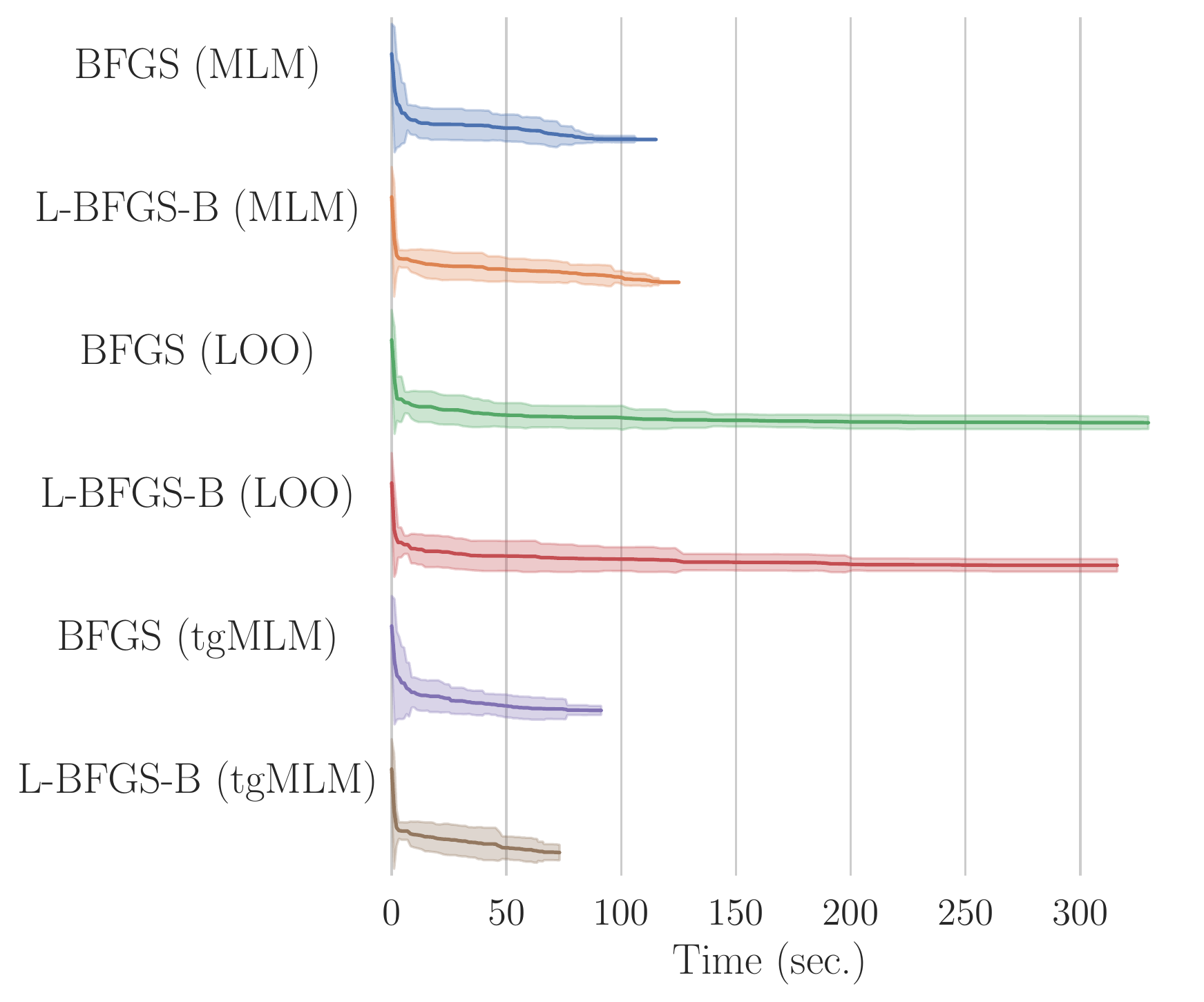}
	}
	\subfigure[Rosenbrock / 100 iters]{
		\includegraphics[width=0.31\linewidth, keepaspectratio]{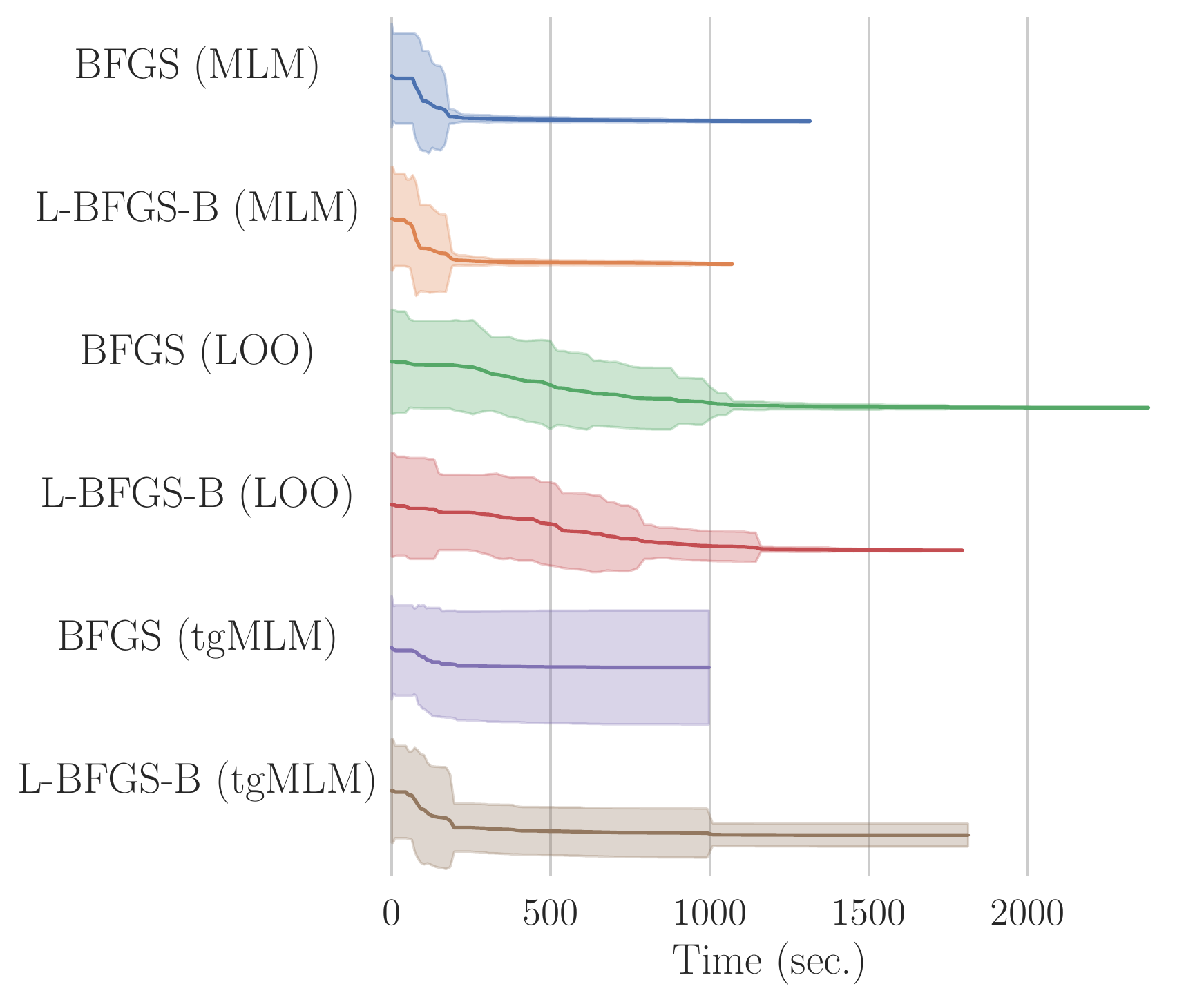}
	}
	\subfigure[Six-hump Camel / 100 iters]{
		\includegraphics[width=0.31\linewidth, keepaspectratio]{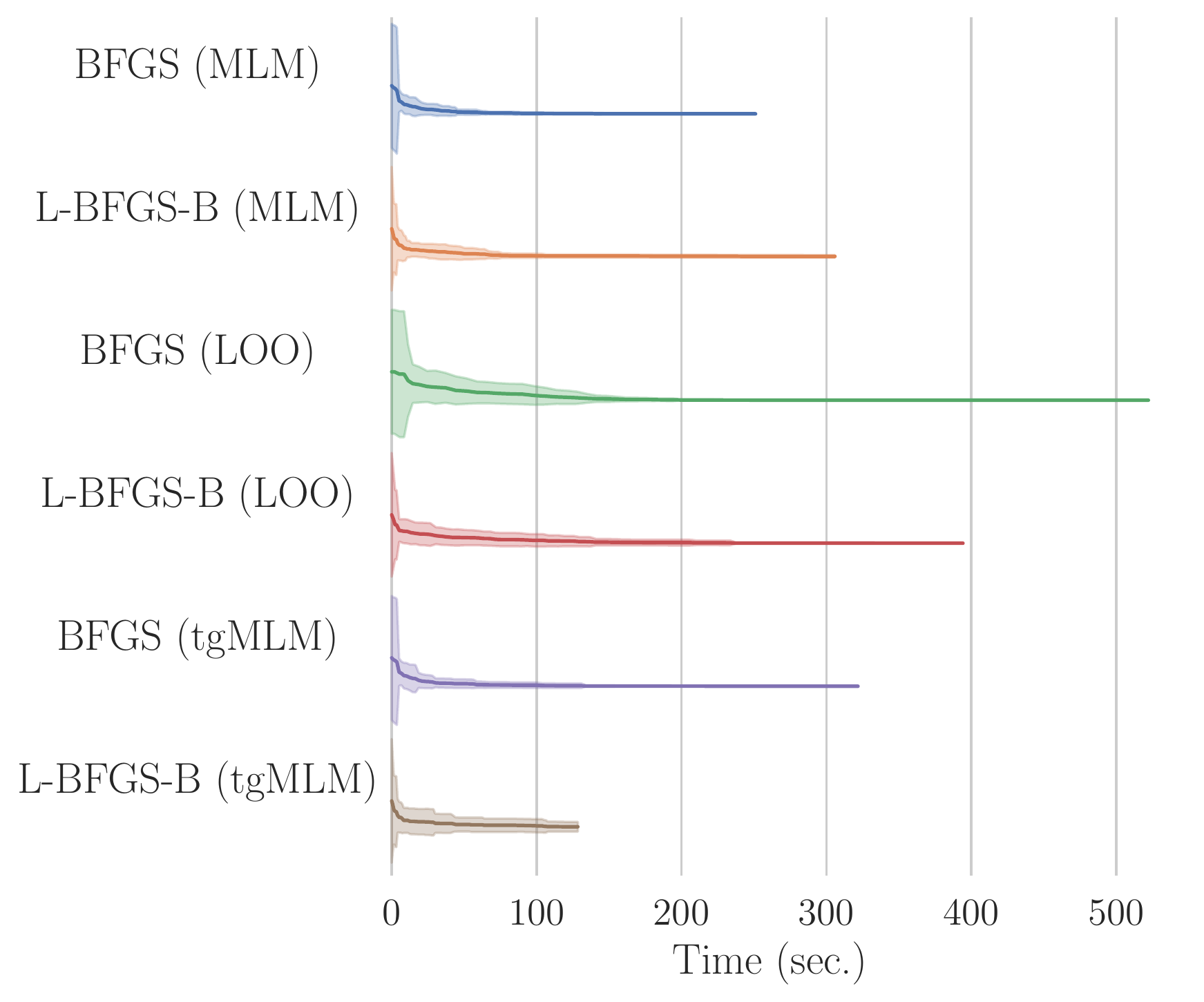}
	}
	\subfigure[RW-1 / 100 iters]{
		\includegraphics[width=0.31\linewidth, keepaspectratio]{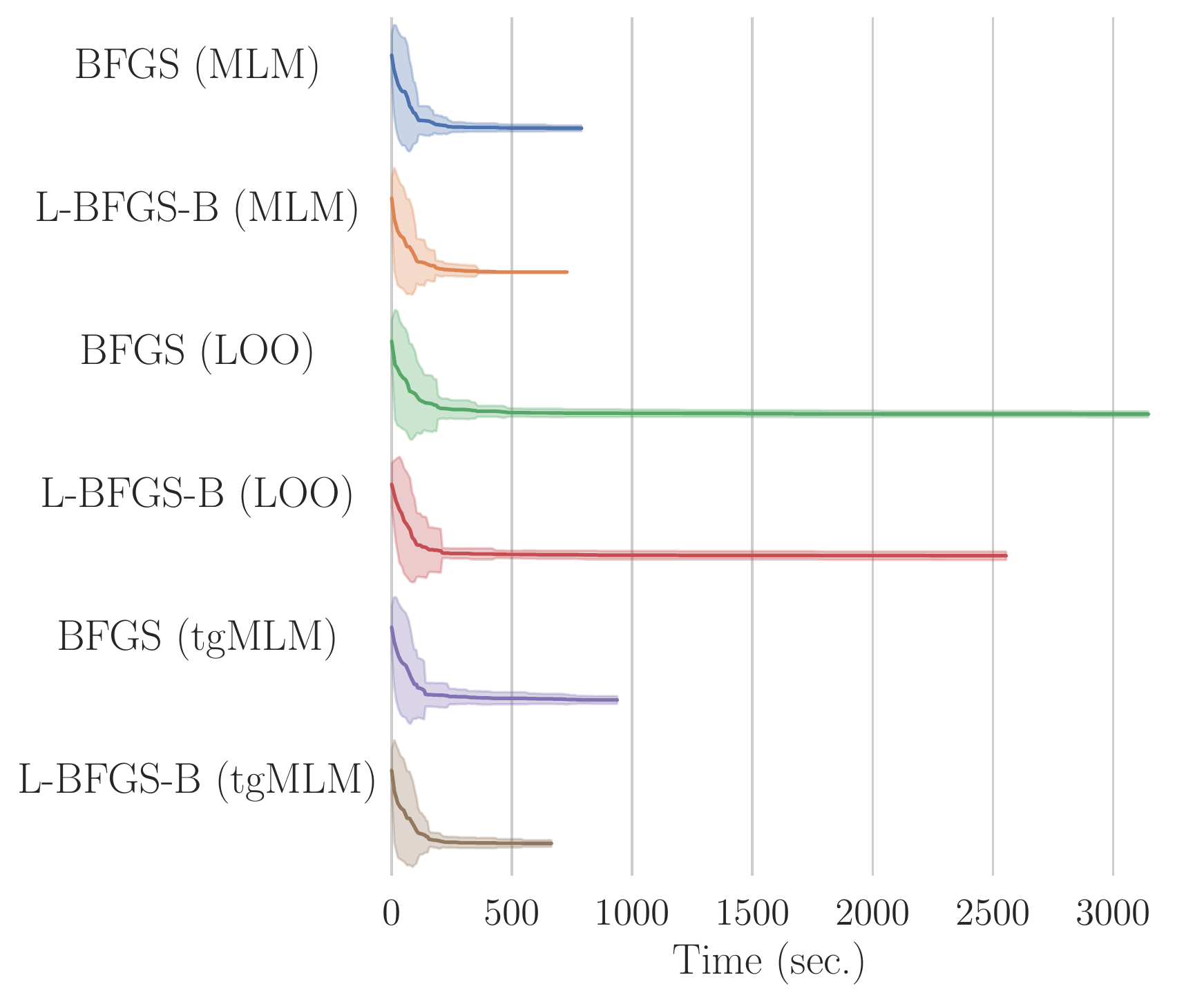}
	}
	\subfigure[RW-2 / 100 iters]{
		\includegraphics[width=0.31\linewidth, keepaspectratio]{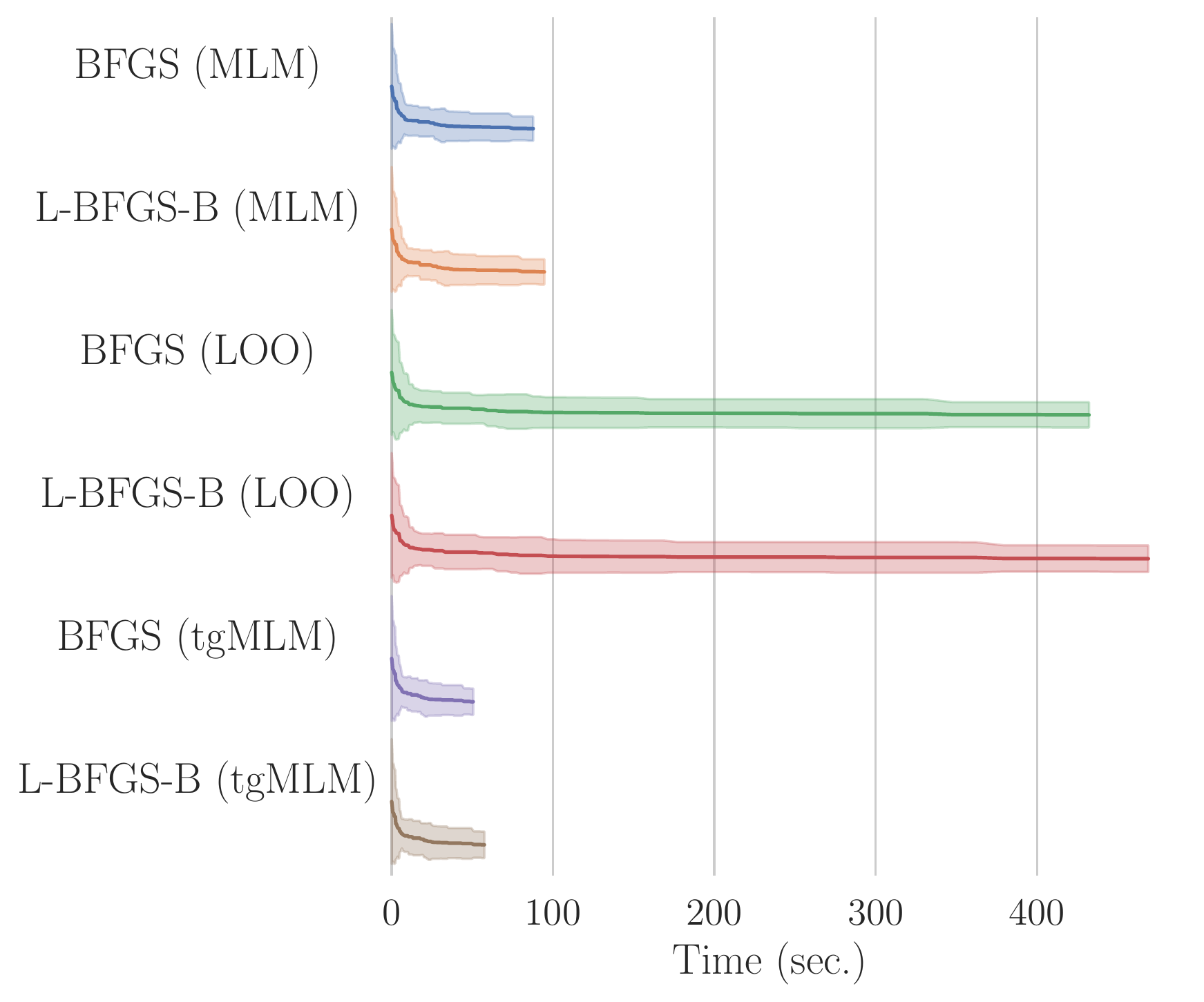}
	}
	\caption{Results on Bayesian optimization of nine benchmark functions and two real-world problems. Horizontal and vertical axes stand for time (seconds) and model selection methods. LOO and tgMLM indicate LOO-CV and threshold-guided MLM, respectively. All experiments are repeated 20 times.}
	\label{fig:all_exps_2}
\end{figure}

We conduct our method on nine benchmark functions and two real-world problems.
The baselines of our method is (i) MLM and (ii) LOO-CV with one of two local optimizers 
(i.e., BFGS and L-BFGS-B).
The free parameter bounds of L-BFGS-B optimizer are $[10^{-2}, 10^{3}]$ for signal strength and $[10^{-2}, 10^{3}]$ for each dimension of length-scales.
Gaussian process regression with Mat\'ern 5/2 kernel and expected improvement criterion are used as surrogate function and acquisition function, respectively.
Acquisition function is optimized by multi-started L-BFGS-B which is started from 100 uniformly sampled points~\citep{KimJ2020ecmlpkdd}.
Furthermore, $\rho$ is set by 5\% of $l_2$ norm of previous free parameter vector 
($\rho = 0.05 \| \blambda_{j - t - 2} \|_2$ using the expressions in Algorithm~\ref{alg:main}),
and $\tau$ is set by the number of iterations which is specified in each caption.
All experiments given three initial points are repeated 20 times.
To implement this work, \texttt{bayeso}~\citep{KimJ2017bayeso} is used, 
and the related codes can be found in 
this repository\footnote{\url{https://github.com/jungtaekkim/practical-bo-with-threshold-guided-mlm}}
and \texttt{bayeso} repository\footnote{\url{https://github.com/jungtaekkim/bayeso}}.

\subsection{Benchmark Functions\label{subsec:bechmarks}}

\begin{table}[t]
	\centering
	\caption{Quantitative results on Bayesian optimization of six benchmark functions. We use same settings and notations in Figure~\ref{fig:all_exps_1} and Figure~\ref{fig:all_exps_2}.}
	\label{tab:benchmark_quant}
	\vspace{10pt}
	\begin{tabular}{cccc}
		\toprule
		& \textbf{Beale} & \textbf{Bohachevsky} & \textbf{Branin} \\
		\midrule
		BFGS (MLM) & $0.244 \pm 0.277$ & $57.139 \pm 52.753$ & $0.052 \pm 0.043$ \\
		L-BFGS-B (MLM) & $0.274 \pm 0.221$ & $50.078 \pm 51.957$ & $0.057 \pm 0.043$ \\
		BFGS (LOO) & $0.171 \pm 0.152$ & $58.017 \pm 56.046$ & $0.012 \pm 0.010$ \\
		L-BFGS-B (LOO) & $0.270 \pm 0.310$ & $51.540 \pm 52.852$ & $0.010 \pm 0.011$ \\
		BFGS (tgMLM) & $0.386 \pm 0.350$ & $56.176 \pm 55.664$ & $0.148 \pm 0.260$ \\
		L-BFGS-B (tgMLM) & $0.304 \pm 0.336$ & $56.176 \pm 55.664$ & $0.013 \pm 0.021$ \\
		\bottomrule
	\end{tabular}
	\vskip 10pt
	\begin{tabular}{cccc}
		\toprule
		& \textbf{Eggholder} & \textbf{Goldstein-Price} & \textbf{Hartmann6D} \\
		\midrule
		BFGS (MLM) & $123.896 \pm 63.258$ & $13.748 \pm 11.916$ & $0.247 \pm 0.100$ \\
		L-BFGS-B (MLM) & $123.896 \pm 63.258$ & $18.681 \pm 15.285$ & $0.213 \pm 0.131$ \\
		BFGS (LOO) & $123.896 \pm 63.258$ & $7.695 \pm 7.190$ & $0.085 \pm 0.054$ \\
		L-BFGS-B (LOO) & $123.896 \pm 63.258$ & $11.096 \pm 10.575$ & $0.831 \pm 0.372$ \\
		BFGS (tgMLM) & $123.896 \pm 63.258$ & $15.064 \pm 16.412$ & $0.251 \pm 0.093$ \\
		L-BFGS-B (tgMLM) & $123.896 \pm 63.258$ & $31.257 \pm 50.050$ & $0.912 \pm 0.373$ \\
		\bottomrule
	\end{tabular}
	\vskip 10pt
	\begin{tabular}{cccc}
		\toprule
		& \textbf{Holder Table} & \textbf{Rosenbrock} & \textbf{Six-hump Camel} \\
		\midrule
		BFGS (MLM) & $0.007 \pm 0.008$ & $41.025 \pm 22.563$ & $0.027 \pm 0.022$ \\
		L-BFGS-B (MLM) & $0.040 \pm 0.054$ & $45.452 \pm 17.158$ & $0.070 \pm 0.059$ \\
		BFGS (LOO) & $0.458 \pm 0.520$ & $26.762 \pm 11.082$ & $0.001 \pm 0.001$ \\
		L-BFGS-B (LOO) & $0.532 \pm 0.518$ & $38.591 \pm 18.493$ & $0.002 \pm 0.003$ \\
		BFGS (tgMLM) & $0.200 \pm 0.373$ & $948.985 \pm 1022.856$ & $0.003 \pm 0.004$ \\
		L-BFGS-B (tgMLM) & $0.350 \pm 0.643$ & $85.426 \pm 205.371$ & $0.220 \pm 0.219$ \\
		\bottomrule
	\end{tabular}
\end{table}

We test nine benchmark functions: Beale, Bohachevsky, Branin, Eggholder, Goldstein-Price, Hartmann6D, Holder Table, Rosenbrock, and Six-hump Camel functions.
Figure~\ref{fig:all_exps_1} and Table~\ref{tab:benchmark_quant} show the results on optimizing such benchmark functions.

\subsection{Real-World Problems\label{subsec:realworld}}

As shown in Figure~\ref{fig:all_exps_2} and Table~\ref{tab:realworld_quant}, 
we test two real-world problems: hyperparameter optimization for (i) classification using random forests (referred to as RW-1)
and (ii) regression with elastic net regularization (referred to as RW-2).

RW-1 trained by the Olivetti face dataset has four hyperparameters to optimize: (i) the number of estimators, 
(ii) maximum depth, (iii) minimum samples to split, and (iv) maximum features used to train a single estimator.
RW-2 trained by the California housing dataset has three hyperparameters: (i) coefficient for $l_1$ regularizer,
(ii) coefficient for $l_2$ regularizer, and (iii) maximum iterations to train.
\texttt{scikit-learn}~\citep{PedregosaF2011jmlr} is used to implement these real-world problems.

\begin{table}[t]
	\centering
	\caption{Quantitative results on Bayesian optimization of real-world problems. All settings follow the settings in Figure~\ref{fig:all_exps_1} and Figure~\ref{fig:all_exps_2}.}
	\label{tab:realworld_quant}
	\vspace{10pt}
	\begin{tabular}{cccc}
		\toprule
		& \textbf{RW-1} & \textbf{RW-2} \\
		\midrule
		BFGS (MLM) & $0.091 \pm 0.007$ & $0.464 \pm 0.024$ \\
		L-BFGS-B (MLM) & $0.087 \pm 0.000$ & $0.463 \pm 0.025$ \\
		BFGS (LOO) & $0.092 \pm 0.007$ & $0.463 \pm 0.025$ \\
		L-BFGS-B (LOO) & $0.099 \pm 0.010$ & $0.460 \pm 0.027$ \\
		BFGS (tgMLM) & $0.093 \pm 0.009$ & $0.460 \pm 0.026$ \\
		L-BFGS-B (tgMLM) & $0.091 \pm 0.008$ & $0.460 \pm 0.027$ \\
		\bottomrule
	\end{tabular}
\end{table}

\section{Future Work and Conclusion\label{sec:future_conclusion}}

In this article, we propose the practical and easy-to-implement Bayesian optimization method.
By our empirical results, we demonstrate our method is practically effective 
in terms of execution time and convergence quality.
However, our algorithm is not sophisticated and theoretical analysis of our method is not provided.
Thus, we can develop our method to more theoretical and more novel method in the future work.
For example, we can bound a discrepancy between Bayesian optimization optimized by MLM and tgMLM with the probability described by some related factors.
Moreover, we can propose a method to select free parameters from historical free parameters.
Since we keep all historical free parameters, those can be used to determine the current free parameters 
without time-consuming model selection procedure.

\bibliography{sjc}
\bibliographystyle{abbrvnat}

\end{document}